\documentclass[lettersize,journal]{IEEEtran}
\usepackage{amsmath,amsfonts}
\usepackage{algorithmic}
\usepackage{algorithm}
\usepackage{array}
\usepackage[caption=false,font=normalsize,labelfont=sf,textfont=sf]{subfig}
\usepackage{textcomp}
\usepackage{stfloats}
\usepackage{url}
\usepackage{verbatim}
\usepackage{graphicx}
\usepackage{cite}
\usepackage{booktabs}
\usepackage{threeparttable}
\usepackage{enumitem}
\usepackage{doi}
\usepackage{stfloats}
\begin{document}

\title{ABFL: Angular Boundary Discontinuity Free Loss \\ for Arbitrary Oriented Object Detection in Aerial Images}

\author{Zifei Zhao, Shengyang Li
\thanks{Z. Zhao and S. Li are with the Key Laboratory of Space Utilization, Technology and Engineering Center for Space Utilization, Chinese Academy of Sciences, Beijing 100094, China, and also with the University of Chinese Academy of Sciences, Beijing 100049, China (e-mail: zhaozifei18@csu.ac.cn; shyli@csu.ac.cn).(Corresponding authors: Shengyang Li.)}
\thanks{Manuscript received April 19, 2021; revised August 16, 2021.}
\thanks{This work has been submitted to the IEEE for possible publication. Copyright may be transferred without notice, after which this version may no longer be accessible.}}

\markboth{Journal of \LaTeX\ Class Files,~Vol.~14, No.~8, August~2021}%
{Shell \MakeLowercase{\textit{et al.}}: A Sample Article Using IEEEtran.cls for IEEE Journals}



\maketitle

\begin{abstract}
Arbitrary oriented object detection (AOOD) in aerial images is a widely concerned and highly challenging task, and plays an important role in many scenarios. The core of AOOD involves the representation, encoding, and feature augmentation of oriented bounding-boxes (Bboxes). Existing methods lack intuitive modeling of angle difference measurement in oriented Bbox representations. Oriented Bboxes under different representations exhibit rotational symmetry with varying periods due to angle periodicity. The angular boundary discontinuity (ABD) problem at periodic boundary positions is caused by rotational symmetry in measuring angular differences. In addition, existing methods also use additional encoding-decoding structures for oriented Bboxes. In this paper, we design an angular boundary free loss (ABFL) based on the von Mises distribution. The ABFL aims to solve the ABD problem when detecting oriented objects. Specifically, ABFL proposes to treat angles as circular data rather than linear data when measuring angle differences, aiming to introduce angle periodicity to alleviate the ABD problem and improve the accuracy of angle difference measurement. In addition, ABFL provides a simple and effective solution for various periodic boundary discontinuities caused by rotational symmetry in AOOD tasks, as it does not require additional encoding-decoding structures for oriented Bboxes. Extensive experiments on the DOTA and HRSC2016 datasets show that the proposed ABFL loss outperforms some state-of-the-art methods focused on addressing the ABD problem.
\end{abstract}


\section{Introduction}
\IEEEPARstart{A}{rbitrary Oriented} Object Detection (AOOD) is a computer vision task that focuses on detecting objects in images and predicting their orientation, particularly when the objects are not aligned with the horizontal or vertical axes. 
AOOD task is useful in aerial scenarios where objects may be oriented at arbitrary angles, such as rescue, reconnaissance, and monitoring \cite{ding2019roitrans,xie2021oriented}. Since aerial images are acquired in nadir observation, the orientation angles of objects in aerial images are arbitrarily oriented in a range of ${0}^{\circ}$ to ${360}^{\circ}$, which increases the difficulty of object detection \cite{thenkabail2018RSHand}.

\begin{figure}[!t]
\centering
\subfloat[Sample of original images in DOTA.]{\includegraphics[width=3.5in]{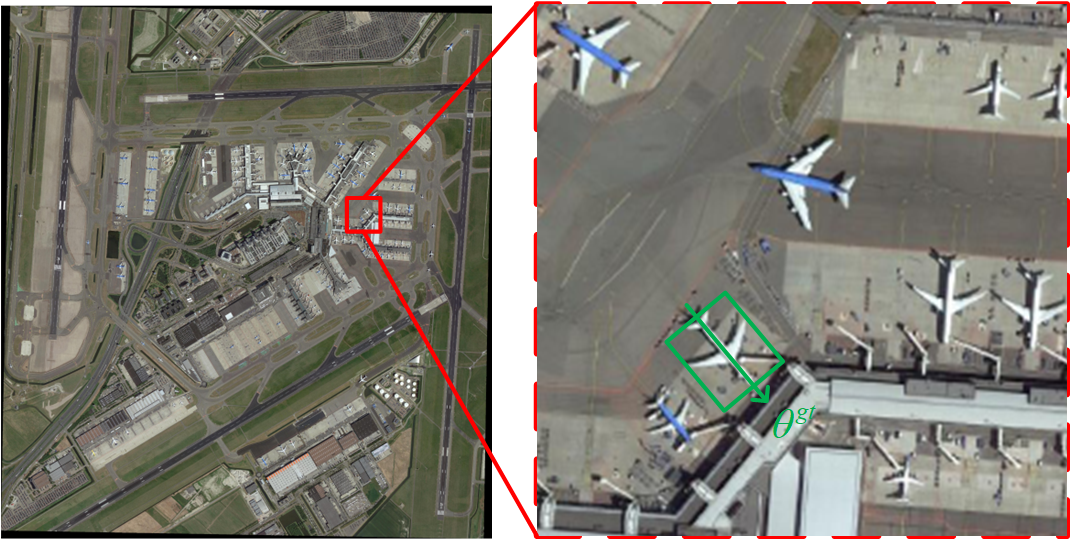}}%
\label{sampleDOTA}
\subfloat[ABD]{\includegraphics[width=1.4in]{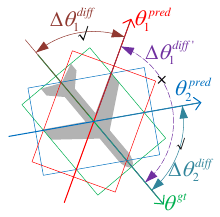}}%
\label{ABDP}
\hfil
\subfloat[ABFL]{\includegraphics[width=2.0in]{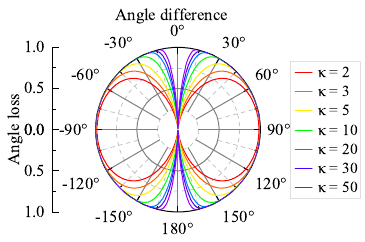}}%
\label{ABFL}
\caption{Angular boundary discontinuity (ABD) problem and angular boundary discontinuity free loss (ABFL). Panel (a) shows a sample of original images in DOTA. Panel (b) shows a case of a plane with its ground-truth(gt) bounding box (in green) and two predicted bounding box samples (in red and blue). ${\theta}^{gt}$ is the oriented angle of the ground-truth bounding box, ${\theta}_1^{pred}$ and ${\theta}_2^{pred}$ denote two predicted oriented angle values, respectively. $\Delta {\theta}_1^{diff}$ and $\Delta {\theta}_2^{diff}$ are the intuitive angle differences between the predicted values and ${\theta}^{gt}$ under the IoU metric. $\Delta {\theta {{_{1}^{diff}}^{'}}}$ is the angle difference between the ${\theta}_1^{pred}$ and ${\theta}^{gt}$ under the linear regression loss, which is inconsistent with the intuitive angle difference $\Delta {\theta}_1^{diff}$. Panel (c) shows our proposed ABFL with various settings in polar coordinates, where angle loss is the radial distance from the origin, and angle difference is the angle from the x-axis (counterclockwise is negative, clockwise is positive). $\kappa>0$ is the concentration factor, which controls the variation of the ABFL over the angle difference.}
\label{fig:proplemsAndABFL} 
\end{figure}

The goal of the AOOD is to detect the object of interest in the images, which is the same as the horizontal object detection task, but differs in processing object orientation. Horizontal object detection algorithms rely on axis-aligned anchor boxes, assuming that objects are aligned with horizontal or vertical axes \cite{ren2015faster,liu2016ssd,redmon2016yolo,law2018cornernet,cai2019cascade,duan2019centernet}. However, this may not accurately represent objects with arbitrary orientations, leading to inaccurate Bbox predictions for such objects. 
AOOD tends to predict the orientation of objects, especially when the objects are not aligned with axes. AOOD methods are mostly modified from horizontal object detectors, using rotated anchor boxes or angle-based representations to accurately capture the boundaries and orientation of objects\cite{ding2019roitrans,xu2020gliding,han2021redet,han2021s2anet,xie2021oriented}. The AOOD methods prefers to focus on the representation, processing, and utilization of additional information about object orientation\cite{ma2018RRPN,yang2019scrdet}. 
There are three types of oriented Bbox representation: adding offset parameters in horizontal boxes representation \cite{ding2019roitrans,WEI2020268midline,xu2020gliding,xie2021oriented}, five-parameter rotated Bbox representation with oriented angle \cite{yang2020FPN-CSL,yang2021DCL}, and encoding conversion representation based on rotated Bbox representations\cite{yang2021GWD,yang2021KLD,yu2023psc}.

Ideally, the most intuitive representation is the five-parameter rotated Bbox representation with oriented angle, which adds a subbranch that specifically predicts the oriented angle of the object\cite{yang2020FPN-CSL,yang2021DCL,yang2021GWD,yang2021KLD,yu2023psc}. However, this solution encounters the angular boundary discontinuity problem, as shown in Fig. \ref{fig:proplemsAndABFL} (b). This problem is caused by the confusion in angular distance metrics due to the periodicity of angular variables. 
The assumption of the existing angular distance metrics is to treat the angular data as linear data, which causes the confusion problem. When the absolute value of the linear variable increases, we assume that it will move away from the origin. Thus, for linear data, 359 is relatively close to 350 and far from the origin (0). However, for angular data, ${359}^{\circ}$ is closer to origin (${0}^{\circ}$) than ${350}^{\circ}$, which is a reflection of the circular data periodicity. To distinguish angular data from the linear data that we are more used to, data of this type is referred to as circular data, which deals with data that can be represented as points on the circumference of the unit circle. 
When the angle is near the boundary position (origin ${0}^{\circ}$) of the unit cycle, there is a large gap between expected values and predicted values. 
This gap leads to the typical phenomenon that for a small angle difference during training, the output value of the loss function is not unique. It cannot be guaranteed that the smaller the angle difference, the smaller the loss value, which directly causes confusion about the network optimization trend. 
This confusion affects the training stability of the network and reduces the accuracy. An ideal loss function should ensure the uniqueness and stability of the angular difference metric.
The research work on ABD problem in aerial image AOOD includes smoothing loss function\cite{yang2019scrdet,qian2021RSDet} and angle encoding conversion\cite{yang2020FPN-CSL,yang2021GWD,yang2021KLD,yu2023psc}.
The latter requires the addition of additional encoding-decoding structures, which increases the complexity of the network. 
Taking account of the circular data‘s periodic nature, we consider that the core of the angular boundary problem is how to design the metric of angular difference for circular data. We propose a loss function called Angular Boundary Free Loss (ABFL), which is specifically designed to handle the angular boundary discontinuity problem in the regression of periodic circular data, as shown in Fig.\ref{fig:proplemsAndABFL} (c).
In summary, this paper has three main contributions:
\begin{itemize}
    \item We propose a novel loss function that can robustly measure the differences of circular variables and address the angular boundary discontinuity problem. 
    \item The proposed loss function does not require an additional encoding decoding structure, which is different from recent angle-regression-based arbitrary oriented object detectors.
    \item The proposed loss function is evaluated on two challenging datasets for AOOD on aerial images,  and ABFL outperforms the methods dedicated to alleviating the angular boundary discontinuity problem.
\end{itemize}

The rest of this paper is organized as follows: 
In section \ref{reltedwork}, we briefly review the related methods of AODD task. 
In section \ref{method}, we introduce the principles and the details of the proposed ABFL loss function. 
In section \ref{experimental}, we describe the experiment results on two datasets to evaluate the performance of ABFL. 
In Section \ref{conclusions}, we summarize the paper and give some prospects.

\section{Related Work}\label{reltedwork}

In this subsection, we mainly investigate related work on arbitrarily oriented object detection and focus on summarizing representative works related to angular boundary discontinuity problem. Readers can refer to \cite{zou2023object} for an exhaustive literature review on object detection.

\subsection{Arbitrary-Oriented Object Detection}
The core problem of arbitrarily oriented object detection is how to enable the detector to achieve fast and robust learning of object orientation information, which is the main difference from horizontal object detection.
The most popular object detectors can be divided into two main categories: two-stage detectors and one-stage detectors.

For two-stage detectors, the network architecture is divided into two stages \cite{ren2015faster,he2017mask,cai2019cascade,sun2021sparsercnn,Cheng2022DOD,Yao2023OIBBR}. In the first stage, various anchors are generated using manually predefined parameters, such as spatial scale and aspect ratio.
Candidate anchors that may contain objects are obtained through foreground background binary classification.
Then, in the second stage, extract the feature of the candidate anchors in the pre-constructed image feature pyramid and predict the representation parameters of the object Bbox.
Works on AOOD extends detectors for horizontal objects to detectors for oriented objects by using rotated Bboxes. AOOD detectors focus more on the use of angle information.
The core of the two-stage method is the design of the rotated anchor generation strategy to achieve more efficient coverage of the object's Bbox and oriented angle. The modeling of orientation information in these works involves adding additional parameters to the axis-aligned Bbox representation and regressing these extended parameters using ln norm loss, such as, RRPN \cite{ma2018RRPN}, RoI transformer \cite{ding2019roitrans}, ReDet \cite{han2021redet}, Oriented RCNN \cite{xie2021oriented}.

For single-stage detectors, also known as the anchor-free method, predict the oriented Bbox directly from the feature map\cite{redmon2016yolo, liu2016ssd,law2018cornernet,duan2019centernet,zhou2019ExtremeNet,han2021s2anet} instead of relying on pre-defined anchor boxes. Some representative works of single-stage detectors are as follows, Fully Convolutional One-Stage Object Detection (FCOS)\cite{tian2019fcos}, which is based on the fully convolutional network (FCN), predicts a Bbox vector and a category vector at each grid in the feature maps. The Bbox vector represents the relative offsets from the center to the four edges of the Bbox. The Bbox vector represents the relative offsets from the center to the four edges of the Bbox. CornerNet\cite{law2018cornernet} detects objects as paired keypoints, which are the top left and bottom right corners of the Bbox. CenterNet\cite{duan2019centernet} detects each object as a triplet including center keypoints and paired corners. ExtremeNet\cite{zhou2019ExtremeNet} predicts four multi-peak heatmaps, each corresponding to one of the four extreme points (top-most, left-most, bottom-most, right-most) of the Bbox. These horizontal object detectors can be conveniently applied to AOOD to predict the orientation by adding an oriented angle prediction branch in the detector's head, commonly termed as rotated FCOS, rotated CenterNet.

Although two-stage detectors can achieve SOTA results in multiple public benchmarks\cite{ding2019roitrans,xie2021oriented,Cheng2023SFRNet}, they also suffer from problems such as slower inference speed. One-stage detectors' performance of model inference speed and the scalability of multi-task modeling are more competitive\cite{han2021s2anet,yang2021GWD,Cheng2022AOPG,yu2023psc}. For the AOOD task, alleviating the angular boundary discontinuity problem is crucial for the performance of single-stage anchor-free detectors.

\begin{figure*}[htb]
\centering
\includegraphics[width=6in]{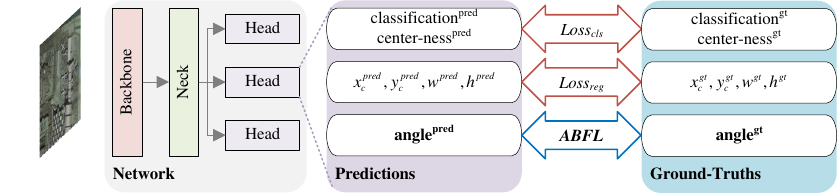}
\caption{The architecture of the Rotated FCOS detector. The architecture includes a backbone, Feature Pyramid Network (FPN), and detection head. The output of the detection head includes the oriented Bbox's categories, coordinates of the center point, width, height, oriented angle, and center-ness. ${{Loss}_{cls}}$, ${{Loss}_{reg}}$, and \textbf{\emph{ABFL}} represent the loss for classification/centerness prediction, Bbox linear regression, and angle circular regression, respectively.}
\label{fig:RFCOSnetwork}
\end{figure*}

\subsection{Angular Boundary Discontinuity}
The angle boundary discontinuity problem is a new challenge faced by angle-regression-detectors, which do not exist in traditional horizontal detectors. 
Recent studies have focused on alleviating the angular boundary discontinuity problem, which can be categorized into four aspects.
\begin{itemize}

\item Smooth loss function. SCRDet \cite{yang2019scrdet} proposes IoU-smooth L1 loss, which concentrates on smoothing outliers in the loss; RSDet \cite{qian2021RSDet} presents a modified version using modulated loss. Both approaches aim to mitigate the problem's effects rather than solving it theoretically.

\item Transforming angular prediction from a regression task to a classification task. Circular smoothing labeling (CSL) \cite{yang2020FPN-CSL} converts angle regression into angle classification to handle the periodicity of oriented angle; Densely Coded Labels (DCL) \cite{yang2021DCL} increases the encoding density of CSL and reduces the number of parameters in the encoding blocks. Gaussian Focal Loss \cite{Wang2022GFL} introduces a dynamic weighting mechanism with Gaussian weight attenuation to achieve accurate angle estimation of oriented objects. While the basic theory of these methods is simple, they exhibit slow convergence, performance sensitivity to hyperparameters, and require complex parameter tuning across different datasets.

\item Converting Oriented Bboxes to Gaussian Distributions. GWD \cite{yang2021GWD} proposes a regression loss based on the Gaussian Wasserstein distance (GWD) by converting the oriented Bbox into a two-dimensional Gaussian distribution. Similarly, The regression loss metric computed in KLD \cite{yang2021KLD} is the Kullback-Leibler Divergence (KLD) between the Gaussian distributions of two oriented Bboxes. While GWD and KLD provide elegant solutions, their predictions are relatively inaccurate, leading to high mAP50 and low mAP75 performance. Additionally, these loss functions exhibit slow convergence during network training and cannot handle the orientation of square-like objects.

\item Encoding angle as vector representations of trigonometric functions. PSC/PSCD \cite{yu2023psc} predicts the orientation angle by converting angles to multiple phase-shifting cosine values, and solves the angle boundary discontinuity problem by leveraging the periodicity of the cosine function. PSC/PSCD introduces an additional complex encoding-decoding module, which converts the angle into a vector composed of trigonometric functions of multiple phases.
\end{itemize}

\section{Method}\label{method}
This section begins with a brief description of the basic network architecture. Next, the concepts related to the angular boundary discontinuity problem are clarified. Then, we present the principles of the Von Mises distribution in detail. Finally, the proposed ABFL loss function based on the Von Mises distribution is introduced.

\subsection{FCOS detector}
Fully Convolutional One-Stage Object Detection(FCOS) \cite{tian2019fcos} is used as the base network, which is an object detector characterized by anchor-free, proposal-free, and dense detection. Rotated FCOS (RFCOS) is a modification of FCOS for arbitrary-oriented object detection (AOOD) tasks. The architecture of RFCOS is shown in Fig.\ref{fig:RFCOSnetwork}. 

The total loss is the weighted summation of the loss functions of each prediction branch, expressed as
\begin{equation}
\label{eq:totalloss}
\begin{aligned}
\boldsymbol{Loss}&={{\omega }_{1}} \cdot \boldsymbol{Los{{s}_{cls}}}+{{\omega }_{2}} \cdot \boldsymbol{Los{{s}_{reg}}}\\
&+{{\omega }_{3}} \cdot \boldsymbol{Los{{s}_{angle}}}+{{\omega }_{4}} \cdot \boldsymbol{Los{{s}_{aux}}}
\end{aligned}
\end{equation}

where $Loss_{cls}$, $Loss_{reg}$, $Loss_{angle}$, and $Loss_{aux}$ are the losses of the classification, Bbox regression, angle regression, and auxiliary branche defined by the detector, respectively. $\omega_{1}$, $\omega_{2}$, $\omega_{3}$ and $\omega_{4}$ are the weight parameters. The auxiliary branche in our work is the center-ness branch defined in FCOS\cite{tian2019fcos}.

\subsection{Oriented Bbox representations}

\begin{figure}[!t]
\centering
\subfloat[OpenCV definition ($RBbox_{oc}$)]{\includegraphics[width=3.6in]{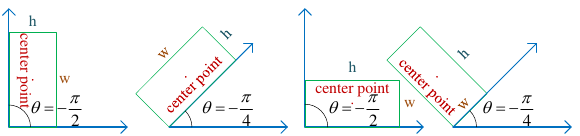}}%
\label{ocdef}
\hfil
\subfloat[long-edge definition ($RBbox_{le}$)]{\includegraphics[width=3.6in]{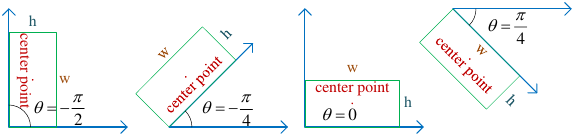}}%
\label{ledef}
\caption{Definitions of oriented Bbox. The center point, w, and h represent the geometric center, width, and height of the oriented Bbox, respectively. $\theta$ represents the oriented angle in the definition.}
\label{fig:RBboxdef} 
\end{figure}

Five parameters $(x, y, w, h, \theta)$ are commonly used to represent the oriented Bbox in AOOD, where $(x, y)$ denotes the center coordinates of the oriented Bbox, $(w, h)$ denotes the width and height of the oriented Bbox, and $\theta$ represents the oriented angle.
There are two common parametric definitions of the oriented Bbox\cite{ma2018RRPN,yang2019scrdet}: the Opencv definition method (denoted by $RBbox_{oc}$) and the long-edge definition method (denoted by $RBbox_{le}$), as shown in Fig.\ref{fig:RBboxdef}.

The main differences between these two definitions are angle range and reference edge. For $RBbox_{le}$ definition, $\theta \in \left[ -\pi/2, \pi/2 \right)$, which denotes the angle between the longer edge ${h}_{le}$ and x-axis, where the angle $\theta$ is negative when ${h}_{le}$ is above the x-axis and, conversely, positive when it is below the x-axis.
When the lengths of the longer and shorter edges are similar, there may be an angle difference of $\pi/2$ between $RBbox_{oc}$ and $RBbox_{le}$, resulting in the square-like problem.
In contrast, for $RBbox_{oc}$, $\theta \in \left[ -\pi/2, 0 \right)$, where the reference edge is the first edge that coincides with the oriented Bbox after counterclockwise rotation of x-axis, which indicates that either the longer edge or the shorter edge in the oriented Bbox may be used as the reference edge, i.e., there is an exchangeability of edges (EoE) problem in $RBbox_{le}$. When the exchange occurs, the angle difference between the two definitions is $\pi/2$.
In previous studies, the detector design is tightly associated with the oriented Bbox definition to avoid specific problems. The $RBbox_{oc}$ is applied to avoid the square-like detection problem \cite{yang2019scrdet,yang2021r3det}, and $RBbox_{le}$ is used to avoid the EoE problem \cite{yang2020FPN-CSL,yang2021DCL}. As an angle-based regression method, ABFL uses $RBbox_{le}$ representation to avoid the EoE problem.

\subsection{Von Mises distribution}
The Von Mises distribution (also known as the circular normal distribution) is a continuous probability distribution for circular data and is often used to model periodic data. It was first proposed by Von Mises in 1918 during his study of the deviation of measured atomic weights from the integral value \cite{mardia2000directional}. This distribution is unimodal and symmetrical about the mean angle difference. This distribution has two parameters: the mean direction and the concentration factor. The mean direction represents the central tendency of the distribution, while the concentration factor represents the degree of clustering around the mean direction, and the distribution has density

\begin{equation}
\label{eq:vmd}
f(\boldsymbol{x}\mid \mu ,\kappa )=\frac{1}{2\pi \cdot {{I}_{p}}(\kappa )}{{\operatorname{e}}^{\kappa \cdot \cos (\boldsymbol{x}-\mu )}}
\end{equation}

where $x$ denotes the oriented angular observation variable, $\mu$ must be a real number and represents the expected value of the oriented angles. ${{I}_{p}}(\kappa)$ is the p-order modified Bessel function, as \eqref{eq:ipk}. 

\begin{equation}
\label{eq:ipk}
{{I}_{p}}(\kappa )=\frac{1}{2\pi }\int_{0}^{2\pi }{\cos (p \cdot \theta) } \cdot {{\operatorname{e}}^{\kappa \cdot \cos \theta }}d\theta
\end{equation}

where $\kappa$ is the concentration factor of the distribution, a real number not less than $0$, describing the dispersion of the data.

\begin{figure}[!t]
\centering
\includegraphics[width=3.4in]{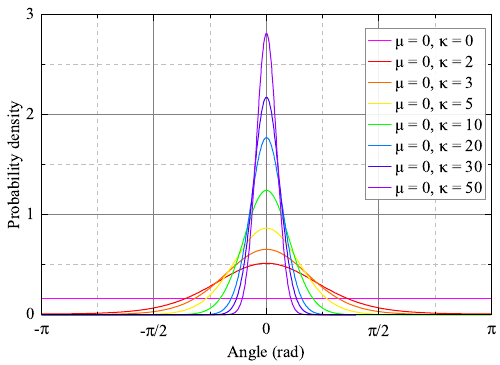}
\caption{Von Mises distribution on x-y plane}
\label{fig:VMD}
\end{figure}

The Von Mises distribution plots corresponding to different $\kappa$ are shown in Fig.\ref {fig:VMD}. When $\kappa=0$, the Von Mises distribution degenerates to a circular uniform distribution $f(x)=\frac{1}{2\pi}$. When $\kappa>0$, $\mu$ denotes the mean direction, and the distribution approximates a Gaussian distribution with mean $\mu$ and variance $\frac{1}{\kappa}$. This distribution is unimodal and reflectively symmetric about $\mu$. When $\kappa \to +\infty$, the distribution tends to a point distribution centered on $\mu$.
The Von Mises distribution has been widely used in various fields such as statistics, physics, and biology due to its ability to model circular data and its mathematical tractability. Its applications include modeling the orientation of fibers in materials, the direction of animal migration, and the direction of wind flow. 
In the above application, the Von Mises distribution of circular data plays a similar role to the normal distribution of linear data.

\subsection{ABFL Loss function}

ABFL is designed to handle the angular boundary discontinuity problem in AOOD. The loss can be expressed as

\begin{equation}
\label{eq:lossangleorigin}
\begin{split}
\boldsymbol{Los{{s}_{angle}}({\theta }_{diff}})&=1-\frac{f(\boldsymbol{{\theta}_{diff}}\mid \mu ,\kappa )}{\gamma}\\
&=1-\frac{{\operatorname{e}}^{\kappa \cdot \cos (\lambda \cdot (\boldsymbol{{\theta}_{diff}}-\mu ))}}{2 \pi \cdot {{I}_{0}}(\kappa ) \cdot \gamma }
\end{split}
\end{equation}
where, ${{\theta }_{diff}}={{\theta }_{pred}}-{{\theta }_{gt}}$, $I_{0}(\kappa)$ is the 0-order modified Bessel function, $\mu =0$. $\kappa \ge 0$ is the concentration factor that controls the distribution density, and $\theta_{diff}$ is the difference between the predicted $\theta_{pred}$ and the ground-truth ${\theta}_{gt}$. $\lambda$ denotes the period adjustment factor, and $\gamma$ represents the normalization factor of the Von Mises distribution density (an approximation of the value of the Von Mises distribution density when $\theta_{diff}$ at $\mu$). $\kappa$ and $\gamma$ are configured in pairs, a series of the pairs tested in this paper is shown in TABLE \ref{tab:kappagammaPairs}.

\begin{table}[h!]
\caption{ the concentration parameter $\kappa$ and the normalization parameter $\gamma$.\label{tab:kappagammaPairs}}
\centering
\begin{tabular}{p{3cm}<{\centering}m{2cm}<{\centering}m{2cm}<{\centering}}
\toprule
Parameter pairs  & $\kappa$ & $\gamma$ \\
\midrule
Pair 1 & 2        & 0.52     \\
Pair 2 & 3        & 0.66     \\
Pair 3 & 5        & 0.87     \\
Pair 4 & 10       & 1.3     \\
Pair 5 & 20       & 1.8     \\
Pair 6 & 30       & 2.2     \\
Pair 7 & 50       & 2.9     \\
\bottomrule
\end{tabular}
\end{table}

\begin{figure}[!t]
\centering
\includegraphics[width=3.5in]{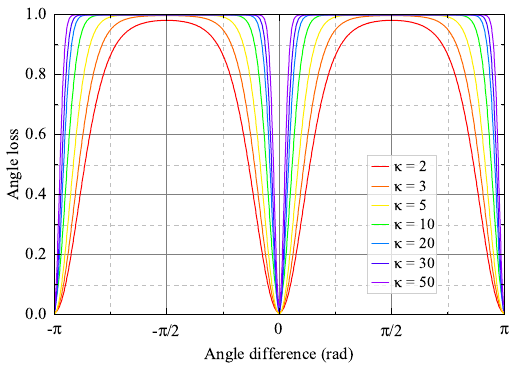}
\caption{ABFL on x-y plane}
\label{fig:VMD_xyplane}
\end{figure}

The period adjustment factor $\lambda$ is related to the cycle of the observed variable. Since the cycle of the Von Mises distribution is $2\pi$, and the cycle of the oriented Bbox in the definition of $RBbox_{le}$ is $\varphi=\pi$. $\lambda$ needs to be set as follows: $\lambda =\frac{2\pi }{\varphi }=2$. The ABFL with different $\kappa$ is shown in Fig.\ref{fig:VMD_xyplane}. 

The Fig.\ref{fig:VMD_xyplane} shows that the angle loss value is 1.0 when the angle difference is $\pi/2$ or $-\pi/2$. The angle loss value is 0 when the angle difference is 0, $\pi$, or $-\pi$. 
ABFL avoids the angular boundary discontinuity problem faced by linear data distance metric and ensures the uniqueness and stability of angle difference measurement.

It should also be mentioned that angle prediction is very unstable at the beginning of training, as the cosine function used in the Von Mises distribution is a periodic function. Therefore, the difference between the angle prediction and the angle ground truth cannot be accurately measured in the loss function, leading to the non-convergence problem of the loss function.

In order to reduce the training difficulty, the output of the angle branch in the detection head needs to be normalized to the range of the orientation angle. Specifically, we design two training strategies to alleviate the non-convergence problem of the loss function.

\textbf{Strategy 1} normalizes the network output values using the torch.atan() function, where the ${\theta }_{pred}$ can be calculated as \eqref{eq:lossangle1} and \eqref{eq:pretheta}.

\begin{equation}
\label{eq:lossangle1}
\boldsymbol{Loss_{angle}}=1-\frac{1}{2\pi \cdot {{I}_{0}}(\kappa ) \cdot \gamma }{{\operatorname{e}}^{\kappa \cdot \cos (2 ({\boldsymbol{\theta }_{pred}}-{\boldsymbol{\theta }_{gt}}))}}
\end{equation}

\begin{equation}
\label{eq:pretheta}
{\boldsymbol{\theta }_{pred}}=atan(\boldsymbol{{X}_{feat}})
\end{equation}

where, ${X}_{feat}$ is the output feature of the last convolution layer in the angle prediction branch of the detection head, and ${\theta }_{pred}$ is the normalized angle prediction with a value domain of $[-\pi/2, \pi/2]$.

\textbf{Strategy 2} modifies the loss function by Forcing the network to learn a predefined range of angles by setting a larger loss value for angle predictions that are not in the defined range, as \eqref{eq:lossangle2}.

\begin{equation}
\label{eq:lossangle2}
\boldsymbol{Los{{s}_{angle}}} = 
\begin{cases}
1-\frac{{{\operatorname{e}}^{\kappa \cdot \cos (2 (\boldsymbol{\theta_{pred}}-\boldsymbol{\theta_{gt}}))}}}{2 \pi \cdot I_{0}(\kappa) \cdot \gamma} &,\,if\left| {\boldsymbol{\theta_{pred}}} \right|\le \frac{\pi }{2}\\ 
\frac{\left| {\boldsymbol{\theta_{pred}}} \right|}{\frac{\pi }{2}} &,\,{\text{otherwise.}} 
\end{cases}
\end{equation}

\subsection{square-like problem}
In order to alleviate the square-like problem that accompanies the long edge definition, we consider adding an aspect ratio threshold (AST) for ABFL to flexibly adapt to objects with different aspect ratios. Specifically, for objects with aspect ratios less than AST, the angular loss is calculated as follows

\begin{equation}
\label{eq:lossangleast}
\begin{split}
\boldsymbol{Loss_{angle}}(\boldsymbol{\theta_{diff}}) = 1-\frac{f(\boldsymbol{\theta_{diff})}}{\gamma}-\frac{f(\boldsymbol{\theta_{diff}}+\frac{\pi }{2})}{\gamma}
\end{split}
\end{equation}
where, $f({\theta }_{diff}))$ denotes the Von Mises distribution with $\lambda=2$, $\mu=0$. ${{\theta }_{diff}}={{\theta }_{pred}}-{{\theta }_{gt}}$.

For objects with small aspect ratios, i.e., objects that are square-like, the angular differences of about $\pm\frac{\pi}{2}$ and $\pm\pi$ have lower loss values. However, for objects with large aspect ratios, only angular differences of about $\pm\pi$ have lower loss values.

\section{Experimental settings}\label{experimental}
In this section, we summarize the experimental results on two typical public datasets, DOTA \cite{ding2021DOTA} and HRSC2016 \cite{Liu2017HRSC2016}, to evaluate the effectiveness of the proposed loss function.

\begin{table}[!t]
  \centering
  \caption{Quantitative comparison between two training strategies for training RFCOS with ABFL (set $\kappa=10$ and $\gamma=1.3$) on DOTA.}
  \label{tab:results_ABFL_traing_strategies}
    \begin{tabular}{p{3.1cm}<{\centering}m{1.3cm}<{\centering}m{1.3cm}<{\centering}m{1.3cm}<{\centering}}
    \toprule
    \textbf{Training Strategies} & \textbf{mAP$_{50}$} & \textbf{mAP$_{75}$} & \textbf{mAP} \\
    \midrule
    \textbf{Strategy 1} & 71.65 & 38.49 & 39.89 \\
    \textbf{Strategy 2} & \textbf{72.12} & \textbf{42.61} & \textbf{41.90} \\
    \bottomrule
    \end{tabular}%
\end{table}%

\begin{table}[!t]
  \centering  \caption{Quantitative comparison of ABFL with various select settings $\kappa$ and $\gamma$ on DOTA.}
  \label{tab:resultsABFLdiffParams}
    \begin{tabular}{p{3.1cm}<{\centering}m{1.3cm}<{\centering}m{1.3cm}<{\centering}m{1.3cm}<{\centering}}
    \toprule
    \textbf{($\kappa$, $\gamma$)} & \textbf{mAP$_{50}$} & \textbf{mAP$_{75}$} & \textbf{mAP} \\
    \midrule
    \textbf{(2, 0.52)} & 71.86 & 40.99 & 40.94 \\
    \textbf{(3, 0.66)} & 71.71 & 41.12 & 41.12 \\
    \textbf{(5, 0.87)} & 71.88 & 41.64 & 41.59 \\
    \textbf{(10, 1.3)} & 72.12 & \textbf{42.61} & 41.90 \\
    \textbf{(20, 1.8)} & 72.28 & 42.48 & \textbf{42.04} \\
    \textbf{(30, 2.2)} & \textbf{72.47} & 42.06 & 41.98 \\
    \textbf{(50, 2.9)} & 72.19 & 41.17 & 41.25 \\
    \bottomrule
    \end{tabular}%
\end{table}%

\begin{table}[!t]
  \centering  \caption{Quantitative comparison of weight for ABFL in total loss on DOTA.}
  \label{tab:resultsABFLdiffWeight}
    \begin{tabular}{p{3.1cm}<{\centering}m{1.3cm}<{\centering}m{1.3cm}<{\centering}m{1.3cm}<{\centering}}
    \toprule
    \textbf{loss weight} & \textbf{mAP$_{50}$} & \textbf{mAP$_{75}$} & \textbf{mAP} \\
    \midrule
    \textbf{0.1} & 72.22 & 42.29 & 41.64 \\
    \textbf{0.2} & 72.12 & \textbf{42.61} & 41.90 \\
    \textbf{0.3} & \textbf{72.27} & 41.97 & 41.86 \\
    \textbf{0.4} & 71.92 & 41.96 & \textbf{41.93} \\
    \textbf{0.5} & 71.95 & 41.31 & 41.86 \\
    \textbf{0.6} & 71.73 & 42.49 & 41.72 \\
    \textbf{0.7} & 71.15 & 41.70 & 41.11 \\
    \textbf{0.8} & 71.04 & 41.22 & 41.19 \\
    \textbf{0.9} & 70.75 & 41.15 & 41.13 \\
    \textbf{1.0} & 70.85 & 41.18 & 40.98 \\
    \bottomrule
    \end{tabular}%
\end{table}%

\begin{table}[!t]
  \centering  \caption{Quantitative comparison of different values of aspect ratio threshold on DOTA.}
  \label{tab:resultsABFLdiffART}
    \begin{tabular}{p{3.1cm}<{\centering}m{1.3cm}<{\centering}m{1.3cm}<{\centering}m{1.3cm}<{\centering}}
    \toprule
    \textbf{Aspect ratio threshold} & \textbf{mAP$_{50}$} & \textbf{mAP$_{75}$} & \textbf{mAP} \\
    \midrule
    \textbf{1.1} & 72.51 & 41.11 & 41.72 \\
    \textbf{1.2} & 71.87 & 41.79 & 41.72 \\
    \textbf{1.3} & \textbf{73.01} & \textbf{42.49} & \textbf{42.26} \\
    \textbf{1.4} & 72.71 & 42.28 & 42.05 \\
    \textbf{1.5} & 72.27 & 42.11 & 42.04 \\
    \bottomrule
    \end{tabular}%
\end{table}%

\begin{table}[!t]
  \centering  \caption{Quantitative comparison of params and GFLOPs on DOTA-v1.0 with input image size of $1024 \times 1024$.}
  \label{tab:resultsABFLeffi}
    \begin{tabular}{p{3cm}m{2cm}<{\centering}m{2cm}<{\centering}}
    \toprule 
    \textbf{Method}                              & \textbf{Params}  & \textbf{GFLOPs} \\
    \midrule
    \textbf{RetinaNet}\cite{lin2017focal}        & 36.5M            & 217 \\
    \textbf{R$^3$Det}\cite{yang2021r3det}        & 42.0M            & 336 \\
    \textbf{Gliding Vertex}\cite{xu2020gliding}  & 41.3M            & 211 \\
    \textbf{RFCOS-CSL}\cite{yang2020FPN-CSL}     & 32.3M            & 216 \\
    \textbf{RFCOS-KLD}\cite{yang2021KLD}         & 31.9M            & 206 \\
    \textbf{RFCOS-PSC}\cite{yu2023psc}           & 31.9M            & 207 \\
    \textbf{RFCOS-ABFL(ours)}                    & 31.8M            & 202 \\
    \bottomrule
    \end{tabular}%
\end{table}%

\begin{figure}[!t]
\centering
\includegraphics[]{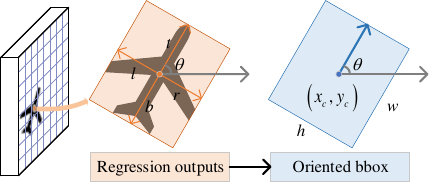}
\caption{The illustration of the transformation between the outputs of the RFCOS regression branch and the oriented Bbox representation.}
\label{fig:outtrans}
\end{figure}

\begin{table}[!t]
    \caption{Results of RFCOS with different loss functions on DOTA.}
    \label{tab:RFCOSwithdifflossonDOTA}
    \centering
    \begin{threeparttable}
    \begin{tabular}{p{3.1cm}m{1.3cm}<{\centering}m{1.3cm}<{\centering}m{1.3cm}<{\centering}}
        \toprule
        \textbf{Method} & $\textbf{mAP}_\textbf{50}$ & $\textbf{mAP}_\textbf{75}$ & \textbf{mAP} \\
        \midrule
        \textbf{RFCOS-Smooth L1} \cite{ren2015faster}     & 71.19  & 37.25  & 39.14  \\
        \textbf{RFCOS-CSL} \cite{yang2020FPN-CSL}     & 70.83  & 38.71  & 39.75  \\
        \textbf{RFCOS-KLD} \cite{yang2021KLD}     & 71.67  & 37.53  & 39.67  \\
        \textbf{RFCOS-PSC} \cite{yu2023psc}     & 71.83  & 39.21  & 40.42  \\
        \textbf{RFCOS-PSCD} \cite{yu2023psc}    & 71.41  & 39.35  & 40.36  \\
        \textbf{RFCOS-SIoU}\tnote{*} \cite{ma2018RRPN} & 71.47  & 39.12  & 40.10  \\
        \textbf{RFCOS-ABFL(ours)}     & \textbf{72.12}  & \textbf{42.61}  & \textbf{41.90}  \\
        \bottomrule
    \end{tabular}
    \begin{tablenotes}
        \footnotesize
        \item[*] Baseline method.
    \end{tablenotes}
    \end{threeparttable}
\end{table}

\begin{figure*}[t]
\centering
\includegraphics[width=7.1in]{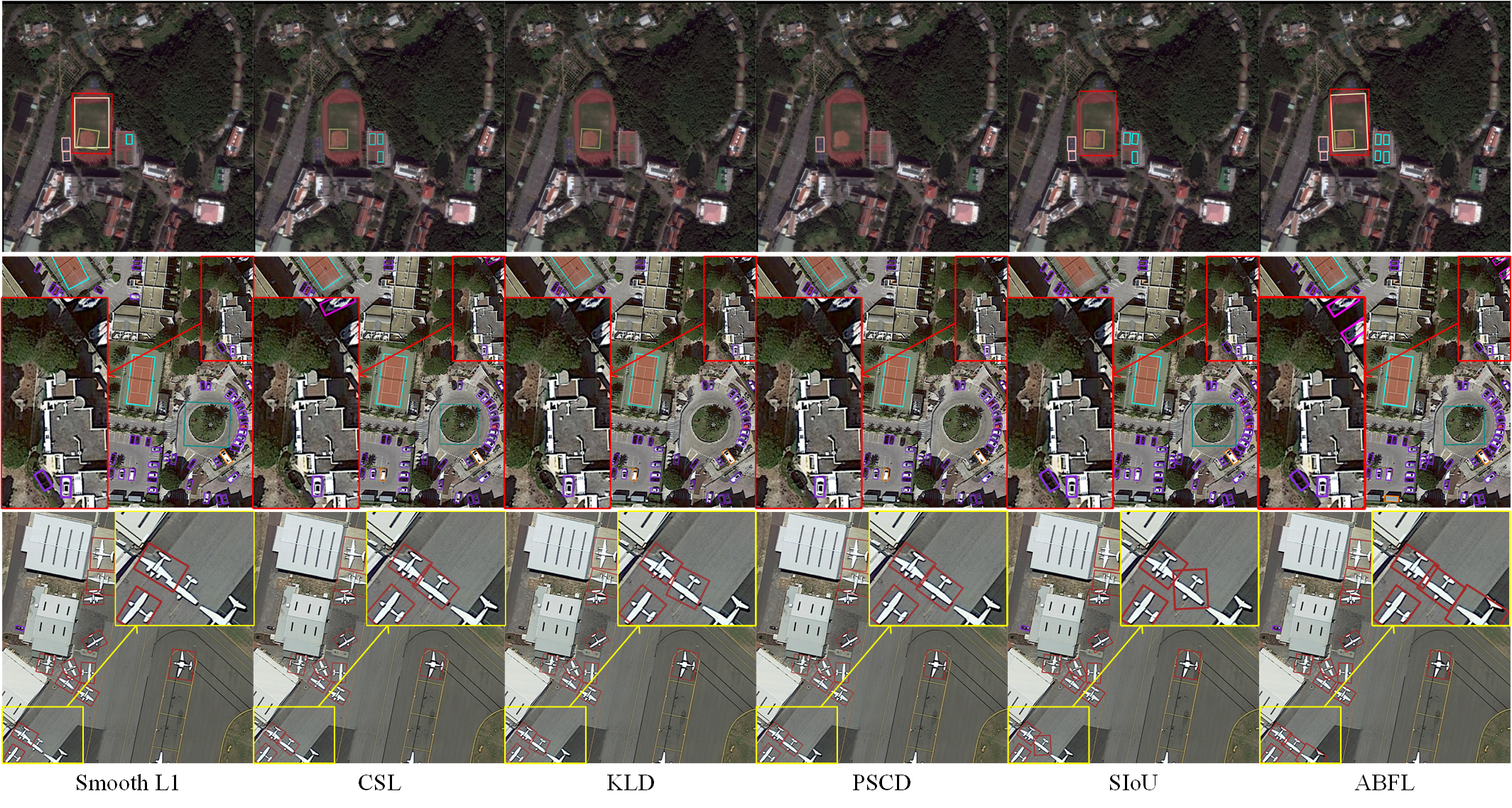}
\includegraphics[width=7.1in]{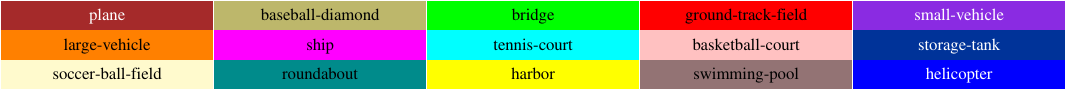}
\caption{Visualization of the RFCOS with different loss functions on DOTA.}
\label{fig:DOTAvisdiffLoss}
\end{figure*}

\begin{table*}[!t]
  \centering
  \footnotesize
  \renewcommand{\arraystretch}{1.43}
  \setlength{\tabcolsep}{1.0mm}
  \setlength{\abovecaptionskip}{2mm}
  \caption{Quantitative comparison between the proposed method and some state-of-the-art methods dedicated to mitigating the discontinuity problem on DOTA. The value in bold font denotes the best performance of each column under single-scale training and testing with ResNet50, and multi-scale training and testing with ResNet50, ResNet101, and ResNet152, respectively.}
  \begin{threeparttable}
    \begin{tabular}{lccccccccccccccccc}
    \toprule
    \textbf{Method} & \textbf{Backbone}\tnote{1} & \textbf{mAP$_{50}$} & \textbf{PL} & \textbf{BD} & \textbf{BR} & \textbf{GTF} & \textbf{SV} & \textbf{LV} & \textbf{SH} & \textbf{TC} & \textbf{BC} & \textbf{ST} & \textbf{SBF} & \textbf{RA} & \textbf{HA} & \textbf{SP} & \textbf{HC} \\
    \midrule
    \multicolumn{18}{c}{Single Scale} \\
    \midrule
    \textbf{RSDet} \cite{qian2021RSDet} & R-50-FPN & 70.79  & \textbf{89.30}  & 82.70  & 47.70  & \textbf{63.90}  & 66.80  & 62.00  & 67.30  & 90.80  & \textbf{85.30}  & 82.40  & \textbf{62.30}  & 62.40  & 65.70  & 68.60  & \textbf{64.60}  \\
    \textbf{FPN-CSL} \cite{yang2020FPN-CSL} & R-50-FPN & 70.92\tnote{2}  & -     & -     & -     & -     & -     & -     & -     & -     & -     & -     & -     & -     & -     & -     & - \\
    \textbf{R3Det-DCL} \cite{yang2021DCL} & R-50-FPN & 71.21\tnote{2}  & -     & -     & -     & -     & -     & -     & -     & -     & -     & -     & -     & -     & -     & -     & - \\
    \textbf{Mask OBB} \cite{wang2019maskobb} & R-50-FPN & 71.43\tnote{2}  & -     & -     & -     & -     & -     & -     & -     & -     & -     & -     & -     & -     & -     & -     & - \\
    \textbf{R3Det-GWD} \cite{yang2021GWD} & R-50-FPN & 71.56\tnote{2}  & -     & -     & -     & -     & -     & -     & -     & -     & -     & -     & -     & -     & -     & -     & - \\
    \textbf{CenterMap} \cite{wang2020CenterMap} & R-50-FPN & 71.74  & 88.88  & 81.24  & \textbf{53.15}  & 60.65  & 78.62  & 66.55  & 78.10  & 88.83  & 77.80  & 83.61  & 49.36  & \textbf{66.19}  & \textbf{72.10}  & \textbf{72.36}  & 58.70  \\
    \textbf{RFCOS-Smooth L1} \cite{ren2015faster} & R-50-FPN & 71.19  & 88.50  & 77.74  & 45.82  & 58.52  & 78.63  & 73.44  & 85.31  & 90.88  & 78.98  & 83.61  & 56.27  & 64.03  & 62.66  & 68.66  & 54.82  \\
    \textbf{RFCOS-CSL} \cite{yang2020FPN-CSL} & R-50-FPN & 70.83  & 88.24  & 74.87  & 41.27  & 61.03  & 79.52  & 78.35  & 87.19  & 90.88  & 81.50  & 84.53  & 54.70  & 62.65  & 62.84  & 68.45  & 46.50  \\
    \textbf{RFCOS-KLD} \cite{yang2021KLD} & R-50-FPN & 71.67  & 89.17  & 75.42  & 49.41  & 56.48  & 79.66  & 76.78  & 86.91  & 90.89  & 83.52  & 84.41  & 58.76  & 62.21  & 63.44  & 67.90  & 50.07  \\
    \textbf{RFCOS-PSC} \cite{yu2023psc} & R-50-FPN & 71.83  & 88.24  & 74.42  & 48.63  & 63.44  & 79.98  & \textbf{80.76}  & \textbf{87.59}  & 90.88  & 82.02  & 71.58  & 59.12  & 60.78  & 65.78  & 71.21  & 53.06  \\
    \textbf{RFCOS-PSCD} \cite{yu2023psc} & R-50-FPN & 71.41  & 88.04  & 73.95  & 48.83  & 63.44  & \textbf{80.01}  & 80.75  & 87.58  & 90.88  & 81.69  & 67.23  & 58.70  & 60.26  & 65.67  & 71.11  & 53.06  \\
    \textbf{RFCOS-SIoU} \cite{ma2018RRPN} & R-50-FPN & 71.47  & 88.57  & 80.69  & 46.23  & 58.35  & 78.48  & 76.04  & 85.90  & 90.90  & 80.57  & 83.03  & 55.94  & 64.74  & 63.44  & 70.21  & 48.88  \\
    \textbf{RFCOS-ABFL} & R-50-FPN & 72.12  & 88.49  & 79.60  & 47.48  & 59.38  & 78.92  & 76.80  & 85.96  & \textbf{90.91}  & 83.31  & 84.15  & 57.15  & 65.73  & 62.84  & 69.70  & 51.63  \\
    \textbf{RFCOS-ABFL w AST} & R-50-FPN & \textbf{73.01}  & 89.05  & \textbf{83.09}  & 48.20  & 63.38  & 79.06  & 77.48  & 86.02  & 90.89  & 82.24  & \textbf{84.60}  & 56.66  & 66.13  & 64.75  & 69.82  & 54.33  \\
    \midrule
    \multicolumn{18}{c}{Multi Scale} \\
    \midrule
    \textbf{Mask OBB} \cite{wang2019maskobb} & R-50-FPN & 75.98  & 89.60  & \textbf{85.82}  & \textbf{56.50}  & 71.18  & 77.62  & 70.45  & 85.04  & 90.18  & 80.10  & 85.30  & 56.60  & 69.43  & 75.45  & 76.71  & 69.70  \\
    \textbf{R3Det-GWD} \cite{yang2021GWD} & R-50-FPN & 77.02  & 89.09  & 84.13  & 55.77  & \textbf{74.48}  & 77.71  & 82.99  & 87.57  & 89.46  & 84.89  & 85.67  & 66.09  & 64.17  & 75.13  & 75.35  & 62.78  \\
    \textbf{RFCOS-SIoU} \cite{ma2018RRPN} & R-50-FPN & 76.80  & \textbf{89.98}  & 82.72  & 54.16  & 69.04  & 80.69  & 83.48  & 87.92  & 90.82  & 81.65  & 84.96  & 61.66  & 71.41  & 74.89  & 77.48  & 61.15  \\
    \textbf{GF-CSL} \cite{Wang2022GFL} & R-50-FPN & 77.54 & 89.77 & 86.15 & 49.85 & 73.01 & 78.56 & 81.77 & 87.95 & 90.84 & \textbf{87.91} & 86.23 & 63.04 & 66.24 & \textbf{77.14} & 79.27 & 65.32 \\
    \textbf{FCOSF} \cite{Rao2023FCOSF} & R-50-FPN & 78.59 & 89.46 & 81.53 & 55.29 & 69.70 & \textbf{81.49} & \textbf{84.79} & \textbf{88.54} & \textbf{90.88} & 87.73 & \textbf{87.09} & \textbf{68.78} & 67.83 & 76.14 & 76.77 & \textbf{72.82} \\
    \textbf{RFCOS-ABFL w AST} & R-50-FPN & \textbf{78.81}  & 88.85  & 83.02  & 56.03  & 70.08  & 81.46  & 84.47  & 88.11  & 90.84  & 84.68  & 86.58  & 67.97  & \textbf{73.19}  & 75.55  & \textbf{82.48}  & 69.86  \\
    \midrule
    \textbf{SCRDet} \cite{yang2019scrdet} & R-101-FPN & 72.61  & 89.98  & 80.65  & 52.09  & 68.36  & 68.36  & 60.32  & 72.41  & 90.85  & \textbf{87.94}  & 86.86  & 65.02  & 66.68  & 66.25  & 68.24  & 65.21  \\
    \textbf{Gliding Vertex} \cite{xu2020gliding} & R-101-FPN & 75.02  & 89.64  & \textbf{85.00}  & 52.26  & \textbf{77.34}  & 73.01  & 73.14  & 86.82  & 90.74  & 79.02  & 86.81  & 59.55  & \textbf{70.91}  & 72.94  & 70.86  & 57.32  \\
    \textbf{R3Det-GWD} \cite{yang2021GWD} & R-101-FPN & 75.66  & 89.64  & 81.70  & 52.52  & 72.96  & 76.02  & 82.60  & 87.17  & 89.57  & 81.25  & 86.09  & 62.24  & 65.74  & 68.05  & 74.96  & 64.38  \\
    \textbf{CenterMap} \cite{wang2020CenterMap} & R-101-FPN & 76.03  & 80.83  & 84.41  & 54.60  & 70.25  & 77.66  & 78.32  & 87.19  & 90.66  & 84.89  & 85.27  & 56.46  & 69.23  & 74.13  & 71.56  & 66.06  \\
    \textbf{SCRDet++} \cite{yang2022scrdet++} & R-101-FPN & 76.81  & \textbf{90.05}  & 84.39  & 55.44  & 73.99  & 77.54  & 71.11  & 86.05  & 90.67  & 87.32  & 87.08  & \textbf{69.62}  & 68.90  & 73.74  & 71.29  & 65.08  \\
    \textbf{R3Det-DCL} \cite{yang2021DCL} & R-101-FPN & 76.97  & 89.14  & 83.93  & 53.05  & 72.55  & 78.13  & 81.97  & 86.94  & 90.36  & 85.98  & 86.94  & 66.19  & 65.66  & 73.72  & 71.53  & \textbf{68.69}  \\
    \textbf{RFCOS-SIoU} \cite{ma2018RRPN} & R-101-FPN & 76.95  & 89.53  & 82.24  & 53.32  & 70.32  & 81.39  & 82.40  & 87.80  & \textbf{90.85}  & 81.92  & 85.27  & 63.57  & 69.26  & 74.64  & 79.87  & 61.81  \\
    \textbf{RFCOS-ABFL w AST} & R-101-FPN & \textbf{78.49}  & 89.65  & 83.29  & \textbf{58.16}  & 67.01  & \textbf{82.03}  & \textbf{83.97}  & \textbf{88.54}  & 90.76  & 85.28  & \textbf{87.32}  & 67.87  & 70.53  & \textbf{77.19}  & \textbf{80.87}  & 64.84  \\
    \midrule
    \textbf{FPN-CSL} \cite{yang2020FPN-CSL} & R-152-FPN & 76.17  & \textbf{90.25}  & \textbf{85.53}  & 54.64  & \textbf{75.31}  & 70.44  & 73.51  & 77.62  & 90.84  & 86.15  & 86.69  & \textbf{69.60}  & 68.04  & 73.83  & 71.10  & 68.93 \\
    \textbf{R3Det-DCL} \cite{yang2021DCL} \tnote{3} & R-152-FPN & 77.37  & 89.26  & 83.60  & 53.54  & 72.76  & 79.04  & 82.56  & 87.31  & 90.67  & \textbf{86.59}  & \textbf{86.98}  & 67.49  & 66.88  & 73.29  & 70.56  & \textbf{69.99}  \\
    \textbf{R3Det-GWD} \cite{yang2021GWD} \tnote{4} & R-152-FPN & 76.18  & 89.55  & 82.28  & 52.39  & 68.30  & 77.86  & 83.40  & 87.48  & 89.56  & 84.27  & 86.14  & 65.38  & 63.25  & 71.33  & 72.36  & 69.21  \\
    \textbf{RFCOS-SIoU} \cite{ma2018RRPN} & R-152-FPN & 76.87  & 88.59  & 82.15  & 51.91  & 68.17  & 81.18  & 83.69  & 88.50  & 90.82  & 83.32  & 86.33  & 60.96  & 68.01  & 74.88  & 79.57  & 64.96  \\
    \textbf{RFCOS-ABFL w AST} \tnote{5} & R-152-FPN & \textbf{78.88}  & 89.73  & 83.85  & \textbf{56.56}  & 70.78  & \textbf{81.85}  & \textbf{84.61}  & \textbf{88.71}  & \textbf{90.85}  & 85.74  & 86.50  & 67.91  & \textbf{71.28}  & \textbf{78.61}  & \textbf{80.69}  & 65.56  \\
    \bottomrule
    \end{tabular}%
    \begin{tablenotes}
        \footnotesize
        \item[1] Column "\textbf{Backbone}" means the feature extraction network, where R-50-FPN, R-101-FPN, and R-152-FPN denote ResNet-50 with FPN, ResNet-101 with FPN, and ResNet-152 with FPN, respectively, FPN denotes Feature Pyramid Network.
        \item[2] Only the $\textbf{mAP}_\textbf{50}$ was reported in these papers.
        \item[3] R3Det-DCL uses an additional feature refinement module and is trained by 40 epochs.
        \item[4] R3Det-GWD uses an additional feature refinement module and is trained by 30 epochs in total with a training and testing scale set to [450, 500, 640, 700, 800, 900, 1,000, 1,100, 1,200].
        \item[5] indicates using NVIDIA TITAN RTX GPU with 24 GB of memory.
    \end{tablenotes}
  \end{threeparttable}
  \label{tab:resultsonDOTA}%
\end{table*}%

\begin{figure*}[!t]
\centering
\includegraphics[width=7.1in]{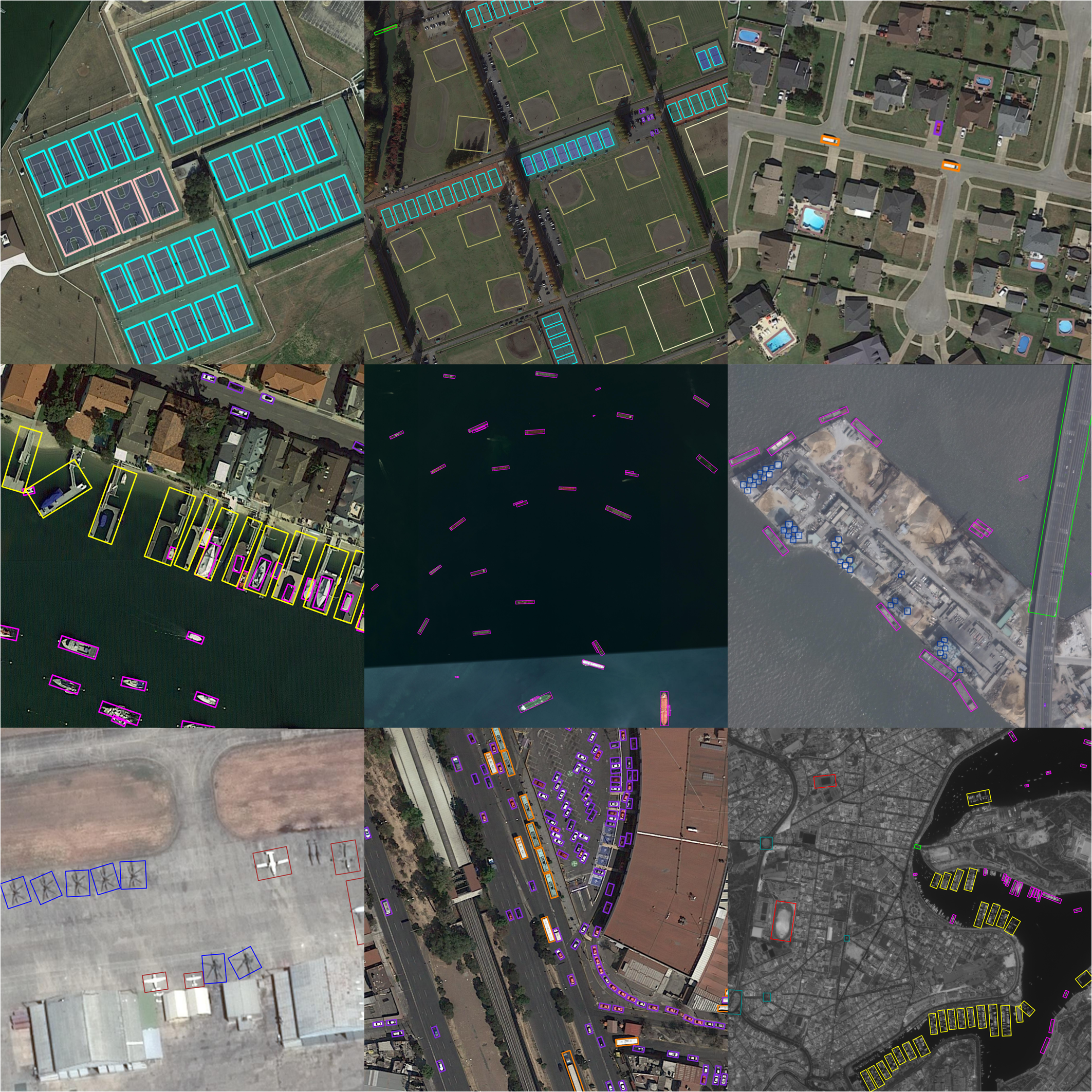}
\includegraphics[width=7.1in]{figures/DOTAvis_label.pdf}
\caption{Visualization of the RFCOS-ABFL on DOTA.}
\label{fig:DOTAresvis}
\end{figure*}

\subsection{Datasets}
DOTA consists of 2,806 large aerial images ranging in size from $800 \times 800$ to $4000 \times 4000$. The dataset is divided into three parts, the train set, the validation set, and the test set containing 1,411, 937, and 458 images respectively. The train and validation sets contain images and labels, while the test set contains images only. DOTA manually defines 15 categories in aerial scenarios with a total of 188,282 instances. The categories include plane (PL), baseball-diamond (BD), bridge (BR), ground-track-field (GTF), small-vehicle (SV), large-vehicle (LV), ship (SH), tennis-court (TC), basketball-court (BC), storage-tank (ST), soccer-ball-field (SBF), roundabout (RA), harbor (HA), swimming-pool (SP) and helicopter (HC). 
For a fair comparison, we follow a standard data processing process. The original image is divided into image patches of size $1024 \times 1024$ with an overlap of 200 pixels. The train set and validation set for model training and test on the test set. After combining the detection results of all patches, we submit the results to the official website to evaluate the accuracy. For multi-scale experiments, the images are resized at three scales $[0.5, 1.0, 1.5]$, the resized images are then cropped into patches of size $1024 \times 1024$ with a stride of 512. 

HRSC2016, as a ship detection dataset, contains various types of ships of arbitrary orientation at sea or near shore. It is commonly used to evaluate arbitrarily oriented object detectors.
The train, validation, and test sets of HRSC2016 include 436, 181, and 444 images, respectively. The image size ranges from $300 \times 300$ to $1500 \times 900$. The train and validation sets are used for training, while the test set is used for testing. In this paper, we scaled the images to $800 \times 800$ for training and testing.

We adopt 50\% random vertical flipping and 50\% random horizontal flipping as data augmentation during single-scale training and use additional random rotation during multi-scale training. And ABFL do not adopt any additional feature enhancement modules.


\subsection{Baselines and Implementation Details}
We trained the network using a single NVIDIA RTX Titan V for 12 epochs with the batchsize of 2 for training. The learning rate is set to 0.002 and divided by 10 at the 8th and the 11th epoch. We optimize the network with SGD, the momentum is 0.9, and the weight decay of 0.0001. The baseline in our work is RFCOS with SkewIoU (SIoU) loss \cite{ma2018RRPN}, which core modification is to replace the IoU loss in FCOS with the SIoU loss. IoU loss is used for horizontal box regression, but SIoU loss is used for rotated box regression. The basic configs of the structure of RFCOS-FPN are the same as the vanilla FCOS with FPN. The backbone network is ResNet-50 \cite{he2016resnet} with FPN \cite{lin2017fpn} (denoted as R-50-FPN). FPN constructs five layers of feature maps, defined as P3 to P7, which are 1/8, 1/16, 1/32, 1/64, and 1/128 of the original image size, respectively. The regression distances for P3 to P7 are (0, 64), (64, 128), (128, 256), (256, 512), and (512, $\infty$), respectively, which allows objects of different sizes assigned to different levels of FPN. If there are still multiple objects assigned to a feature grid, we choose the gt with the smallest area as the object among the multiple gts. Rotated FCOS infers the oriented Bbox predictions pixel by pixel in the feature grid of each layer of the five FPN feature maps. The prediction is a five-dimension vector $(t,b,l,r,\theta)$ that represents the four distances and the oriented angle. 
The illustration of the transform between this vector and the representation of the oriented Bbox $(x_{c},y_{c},w,h,\theta)$ is shown in Fig.\ref{fig:outtrans}. In total loss, $\omega_{1}$, $\omega_{2}$ and $\omega_{4}$ are set to 1.0, $\omega_{3}$ is set to 0.2 by default.

The detection accuracy is measured by the Intersection of Union (IoU) between $\text{BBoxes}_\text{pred}$ and $\text{BBoxes}_\text{gt}$. For DOTA dataset, we list the mAP$_{50}$, mAP$_{75}$, mAP values under COCO metrics\cite{lin2014coco}. mAP$_{50}$, mAP$_{75}$, mAP refers to mean Average Precision (mAP) at IoU=0.5, at IoU=0.75, and at IoU=0.50:0.05:0.95, respectively. The values of IoU indicate the threshold value for determining whether an object is detected or not. IoU=0.50 means that if the IoU is not less than 0.50, the detection is judged to be successful, which is the most commonly used metric. IoU=0.75 is the strict metric that the detection is successful when the IoU is greater than or equal to 0.75. IoU=.50:.05:.95 means that the threshold value is taken from 0.50 to 0.95 at a stride of 0.05, and then calculate the mean value, which is the most comprehensive metric.
On the HRSC2016 dataset, we list the mAP(07) values under PASCAL VOC 2007 metrics\cite{everingham2009pascal}.

\subsection{Ablation studies}

\begin{itemize}
\item Ablation studies of training strategies

The quantitative comparison of proposed training strategies to avoid the non-convergence problem of the ABFL at the beginning of training is shown in TABLE \ref{tab:results_ABFL_traing_strategies}. Strategy II shows a significant accuracy advantage, mAP$_{50}$ is improved by 0.63\%. In particular, for the strict metric, mAP$_{75}$ can be improved by nearly 4 points. So we choose strategy 2 in the subsequent experiments.

\item Ablation studies of hyper-parameters ($\kappa$, $\gamma$)

In most existing methods\cite{yang2020FPN-CSL,yang2021GWD,yang2021KLD}, hyper-parameters may seriously affect performance. The optimal parameters are different in different scenarios and datasets, and they often require costly tuning. The manually adjustable hyper-parameters of ABFL are $\kappa$ and $\gamma$, which need to be set in pairs. We evaluate several pairs of the parameters, and the results are shown in TABLE \ref{tab:resultsABFLdiffParams}. According to the results, the impact of $\kappa$ and $\gamma$ parameters is quite limited when $\kappa \in [10, 50]$. Overall, with more consideration for the comprehensive metric mAP$_{75}$ and mAP, we recommend setting $\kappa$ to 10, and $\gamma$ to 1.3 as the basic parameter pairs.

\item Choice of the weight for ABFL in the total loss function

As shown in the TABLE \ref{tab:resultsABFLdiffWeight}, we test different loss weights for ABFL. When the weight value is 0.2, the result is optimal. Therefore, the weight of ABFL is set as 0.2 in all experiments.

\item Effects of the aspect ratio threshold

As shown in the TABLE \ref{tab:resultsABFLdiffART}, 
the aspect ratio threshold is used to alleviate the square-like problem in the long-edge definition, which is similar to the trick of previous works\cite{yang2021DCL,Wang2022GFL}.
As shown in the table, the optimal result is achieved when the aspect ratio threshold is set as 1.3.

\item Computation efficiency of ABFL

We compare the parameter quantity and computational complexity of some methods on DOTA-v1.0. All methods adopt R-50-FPN as the backbone. The size of the input image is $1024 \times 1024$. As shown in Table \ref{tab:resultsABFLeffi}, we can conclude that ABFL is an effective detector with a lower parameter quantity (31.8M) and computational complexity (202 GFLOPs). 

\end{itemize}

\subsection{Comparison with some state-of-the-art methods}

We compare ABFL with some state-of-the-art methods dedicated to mitigating the angular boundary discontinuity problem, and the quantitative results on the DOTA, HRSC dataset are shown in TABLEs \ref{tab:RFCOSwithdifflossonDOTA}, \ref{tab:resultsonDOTA}, \ref{tab:resultsonHRSC}. 

\subsubsection{Results of RFCOS with different loss functions on DOTA}
We compare the ABFL with some losses dedicated to mitigating the angular boundary discontinuity problem, and the results are shown in TABLE \ref{tab:RFCOSwithdifflossonDOTA}. For mAP$_{50}$, ABFL achieves a 0.29\% AP gap with the closest loss function. The performance of ABFL is on average 1.29 and 3.9 points higher than CSL, with an angle classification branch in the detect head, in mAP$_{50}$ and mAP$_{75}$, respectively. 
For the strict metric mAP$_{75}$, ABFL can be improved by 3.26-5.08 points AP. For the most comprehensive metric mAP, ABFL achieves 41.9\%. 
ABFL is comparable to PSC in the mAP$_{50}$ metric but is on the high side in mAP$_{75}$. The visualization result is shown in Fig.\ref{fig:DOTAvisdiffLoss}. 
In the first line, ABFL detects more objects and accurately predicts their orientation angle. 
In the second row, ABFL successfully predicts two ships and two small vehicles in the upper right corner.
In the third row, ABFL can still accurately predict objects with partial overlap or cropping.

\subsubsection{Results on DOTA}

As shown in Table \ref{tab:resultsonDOTA}, we compare the performance with single-scale training. RFCOS with ABFL achieves 72.14\% mAP, outperforming the other methods. 

We also compared the results of different backbones with multi-scale training. The mAP of ABFL using ResNet-50 with FPN improved by 5.44\%, and also achieves the best performance with R-101-FPN and R-152-FPN backbone. It should be noted that R3Det-GWD is trained by 30 epochs in total with 9 image pyramid scales, and using additional feature alignment modules \cite{yang2021GWD}. 
We visualize diverse samples containing objects with various scales, complex backgrounds, and diverse orientations. As shown in Fig.\ref{fig:DOTAresvis}, it can be observed that ABFL enables accurate modeling of object orientation angle. For several categories such as vehicle, harbor, and boat, ABFL can accurately predict orientated angles. However, for the helicopter, the oriented angle prediction is poor. We suggest that there are three reasons for this: firstly, it is too rare in the dataset. Secondly, its main orientation information is not obvious. And thirdly, it is affected by the square-like problem due to its small aspect ratio.

\subsubsection{Results on HRSC2016}
\begin{figure}[!]
\centering
\includegraphics[width=3.5in]{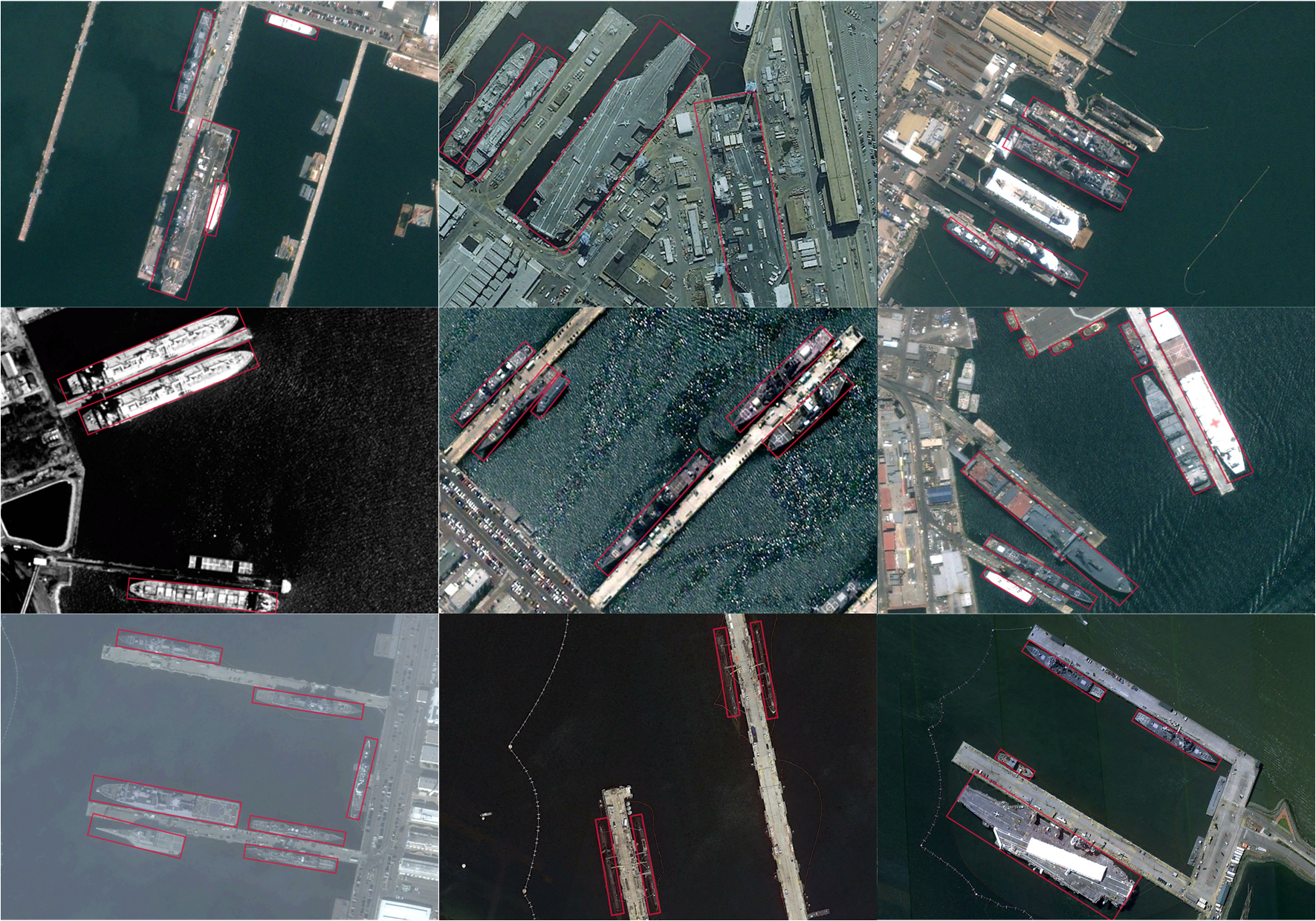}
\caption{Visualization of the RFCOS-ABFL on HRSC2016.}
\label{fig:HRSCresvis}
\end{figure}

Table \ref{tab:resultsonHRSC} shows that RFCOS with ABFL achieves a competitive performance: 89.98\%/90.30\% in terms of the PASCAL VOC 2007 evaluation metric on Resnet50 with FPN and Resnet101 with FPN, respectively. The visualization is shown in Fig\ref{fig:HRSCresvis}. ABFL can accurately predict the ship's oriented angle.


\begin{table}[!t]
  \centering
  \caption{Quantitative comparison between the proposed method and some state-of-the-art methods on HRSC2016.}
  \begin{threeparttable}
    \begin{tabular}{p{3cm}m{2cm}<{\centering}m{2cm}<{\centering}}
    \toprule
    \textbf{Method} & \multicolumn{1}{c}{\textbf{Backbone}} & \textbf{mAP(07)}\tnote{*} \\
    \midrule
    \textbf{RSDet} \cite{qian2021RSDet} & R-50-FPN & 86.50  \\
    \textbf{RIDet} \cite{ming2021RIDet} & R-50-FPN & 89.47  \\
    \textbf{RFCOS-KLD} \cite{yang2021KLD} & R-50-FPN & 89.76  \\
    \textbf{RFCOS-CSL} \cite{yang2020FPN-CSL} & R-50-FPN & 89.84  \\
    \textbf{RFCOS-PSC} \cite{yu2023psc} & R-50-FPN & 90.06  \\
    \textbf{RFCOS-PSCD} \cite{yu2023psc} & R-50-FPN & 89.91  \\
    \textbf{RFCOS-SIoU*} \cite{ma2018RRPN} & R-50-FPN & 89.51  \\
    \textbf{RFCOS-ABFL(ours)} & R-50-FPN & \textbf{89.98}  \\
    \midrule
    \textbf{Rotated RPN} \cite{ma2018RRPN} & R-101 & 79.08  \\
    \textbf{RoI Transformer} \cite{ding2019roitrans} & R-101-FPN & 86.20  \\
    \textbf{Gliding Vertex} \cite{xu2020gliding} & R-101-FPN & 88.20  \\
    \textbf{OBD} \cite{liu2021OBD} & R-101-FPN & 89.22  \\
    \textbf{R3Det-DCL} \cite{yang2021DCL} & R-101-FPN & 89.46  \\
    \textbf{FPN-CSL} \cite{yang2020FPN-CSL} & R-101-FPN & 89.62  \\
    \textbf{RIDet} \cite{ming2021RIDet} & R-101-FPN & 89.63  \\
    \textbf{R3Det-GWD} \cite{yang2021GWD} & R-101-FPN & 89.85  \\
    \textbf{S$^2$A-Net} \cite{han2021s2anet} & R-101-FPN & 90.00  \\
    \textbf{RFCOS-ABFL(ours)} & R-101-FPN & \textbf{90.30}  \\
    \bottomrule
    \end{tabular}%
    \begin{tablenotes}
        \footnotesize
        \item[*] \textbf{mAP(07)} denotes PASCAL VOC 2007 metric.
    \end{tablenotes}
  \end{threeparttable}
  \label{tab:resultsonHRSC}%
\end{table}%

\section{Conclusions}\label{conclusions}
In this paper, we present a novel angular boundary free loss (ABFL) for evaluating the differences of periodic variables. The conclusions are summarized as follows:
\begin{itemize}

\item ABFL solves the angular boundary discontinuity problem in AOOD task and achieves accurate measurement of angle differences.

\item ABFL is simple and highly effective and does not require any additional encoding-decoding module to represent the oriented angle. Its advantage is to reduce the complexity of oriented Bbox representation without affecting the model inference speed.

\item The disadvantage of ABFL is that its effectiveness still needs to be improved when detecting objects with small aspect ratios and fewer samples.

\end{itemize}

Furthermore, ABFL can be used for oriented object detection in many scenarios, such as detecting rotational symmetry with different periods and distinguishing specific azimuths, which will be the motivation for our future work.


{
\small
\bibliographystyle{IEEEtran}
\bibliography{ABFL}

\begin{thebibliography}{10}
\providecommand{\url}[1]{#1}
\csname url@samestyle\endcsname
\providecommand{\newblock}{\relax}
\providecommand{\bibinfo}[2]{#2}
\providecommand{\BIBentrySTDinterwordspacing}{\spaceskip=0pt\relax}
\providecommand{\BIBentryALTinterwordstretchfactor}{4}
\providecommand{\BIBentryALTinterwordspacing}{\spaceskip=\fontdimen2\font plus
\BIBentryALTinterwordstretchfactor\fontdimen3\font minus \fontdimen4\font\relax}
\providecommand{\BIBforeignlanguage}[2]{{%
\expandafter\ifx\csname l@#1\endcsname\relax
\typeout{** WARNING: IEEEtran.bst: No hyphenation pattern has been}%
\typeout{** loaded for the language `#1'. Using the pattern for}%
\typeout{** the default language instead.}%
\else
\language=\csname l@#1\endcsname
\fi
#2}}
\providecommand{\BIBdecl}{\relax}
\BIBdecl

\bibitem{ding2019roitrans}
J.~Ding, N.~Xue, Y.~Long, G.-S. Xia, and Q.~Lu, ``Learning roi transformer for oriented object detection in aerial images,'' in \emph{Proceedings of the IEEE/CVF Conference on Computer Vision and Pattern Recognition}, 2019, pp. 2849--2858.

\bibitem{xie2021oriented}
X.~Xie, G.~Cheng, J.~Wang, X.~Yao, and J.~Han, ``Oriented r-cnn for object detection,'' in \emph{Proceedings of the IEEE/CVF International Conference on Computer Vision}, 2021, pp. 3520--3529.

\bibitem{thenkabail2018RSHand}
P.~Thenkabail, \emph{Remote Sensing Handbook-Three Volume Set}.\hskip 1em plus 0.5em minus 0.4em\relax CRC Press, 2018.

\bibitem{ren2015faster}
S.~Ren, K.~He, R.~Girshick, and J.~Sun, ``Faster r-cnn: Towards real-time object detection with region proposal networks,'' \emph{Advances in neural information processing systems}, vol.~28, 2015.

\bibitem{liu2016ssd}
W.~Liu, D.~Anguelov, D.~Erhan, C.~Szegedy, S.~Reed, C.-Y. Fu, and A.~C. Berg, ``Ssd: Single shot multibox detector,'' in \emph{Computer Vision--ECCV 2016: 14th European Conference, Amsterdam, The Netherlands, October 11--14, 2016, Proceedings, Part I 14}.\hskip 1em plus 0.5em minus 0.4em\relax Springer, 2016, pp. 21--37.

\bibitem{redmon2016yolo}
J.~Redmon, S.~Divvala, R.~Girshick, and A.~Farhadi, ``You only look once: Unified, real-time object detection,'' in \emph{Proceedings of the IEEE conference on computer vision and pattern recognition}, 2016, pp. 779--788.

\bibitem{law2018cornernet}
H.~Law and J.~Deng, ``Cornernet: Detecting objects as paired keypoints,'' in \emph{Proceedings of the European conference on computer vision (ECCV)}, 2018, pp. 734--750.

\bibitem{cai2019cascade}
Z.~Cai and N.~Vasconcelos, ``Cascade r-cnn: high quality object detection and instance segmentation,'' \emph{IEEE transactions on pattern analysis and machine intelligence}, vol.~43, no.~5, pp. 1483--1498, 2019.

\bibitem{duan2019centernet}
K.~Duan, S.~Bai, L.~Xie, H.~Qi, Q.~Huang, and Q.~Tian, ``Centernet: Keypoint triplets for object detection,'' in \emph{Proceedings of the IEEE/CVF international conference on computer vision}, 2019, pp. 6569--6578.

\bibitem{xu2020gliding}
Y.~Xu, M.~Fu, Q.~Wang, Y.~Wang, K.~Chen, G.-S. Xia, and X.~Bai, ``Gliding vertex on the horizontal bounding box for multi-oriented object detection,'' \emph{IEEE transactions on pattern analysis and machine intelligence}, vol.~43, no.~4, pp. 1452--1459, 2020.

\bibitem{han2021redet}
J.~Han, J.~Ding, N.~Xue, and G.-S. Xia, ``Redet: A rotation-equivariant detector for aerial object detection,'' in \emph{Proceedings of the IEEE/CVF Conference on Computer Vision and Pattern Recognition}, 2021, pp. 2786--2795.

\bibitem{han2021s2anet}
J.~Han, J.~Ding, J.~Li, and G.-S. Xia, ``Align deep features for oriented object detection,'' \emph{IEEE Transactions on Geoscience and Remote Sensing}, vol.~60, pp. 1--11, 2021.

\bibitem{ma2018RRPN}
J.~Ma, W.~Shao, H.~Ye, L.~Wang, H.~Wang, Y.~Zheng, and X.~Xue, ``Arbitrary-oriented scene text detection via rotation proposals,'' \emph{IEEE transactions on multimedia}, vol.~20, no.~11, pp. 3111--3122, 2018.

\bibitem{yang2019scrdet}
X.~Yang, J.~Yang, J.~Yan, Y.~Zhang, T.~Zhang, Z.~Guo, X.~Sun, and K.~Fu, ``Scrdet: Towards more robust detection for small, cluttered and rotated objects,'' in \emph{Proceedings of the IEEE/CVF International Conference on Computer Vision}, 2019, pp. 8232--8241.

\bibitem{WEI2020268midline}
H.~Wei, Y.~Zhang, Z.~Chang, H.~Li, H.~Wang, and X.~Sun, ``Oriented objects as pairs of middle lines,'' \emph{ISPRS Journal of Photogrammetry and Remote Sensing}, vol. 169, pp. 268--279, 2020.

\bibitem{yang2020FPN-CSL}
X.~Yang and J.~Yan, ``Arbitrary-oriented object detection with circular smooth label,'' in \emph{Computer Vision--ECCV 2020: 16th European Conference, Glasgow, UK, August 23--28, 2020, Proceedings, Part VIII 16}.\hskip 1em plus 0.5em minus 0.4em\relax Springer, 2020, pp. 677--694.

\bibitem{yang2021DCL}
X.~Yang, L.~Hou, Y.~Zhou, W.~Wang, and J.~Yan, ``Dense label encoding for boundary discontinuity free rotation detection,'' in \emph{Proceedings of the IEEE/CVF conference on computer vision and pattern recognition}, 2021, pp. 15\,819--15\,829.

\bibitem{yang2021GWD}
X.~Yang, J.~Yan, Q.~Ming, W.~Wang, X.~Zhang, and Q.~Tian, ``Rethinking rotated object detection with gaussian wasserstein distance loss,'' in \emph{International Conference on Machine Learning}.\hskip 1em plus 0.5em minus 0.4em\relax PMLR, 2021, pp. 11\,830--11\,841.

\bibitem{yang2021KLD}
X.~Yang, X.~Yang, J.~Yang, Q.~Ming, W.~Wang, Q.~Tian, and J.~Yan, ``Learning high-precision bounding box for rotated object detection via kullback-leibler divergence,'' \emph{Advances in Neural Information Processing Systems}, vol.~34, pp. 18\,381--18\,394, 2021.

\bibitem{yu2023psc}
Y.~Yu and F.~Da, ``Phase-shifting coder: Predicting accurate orientation in oriented object detection,'' in \emph{Proceedings of the IEEE/CVF Conference on Computer Vision and Pattern Recognition}, 2023, pp. 13\,354--13\,363.

\bibitem{qian2021RSDet}
W.~Qian, X.~Yang, S.~Peng, J.~Yan, and Y.~Guo, ``Learning modulated loss for rotated object detection,'' in \emph{Proceedings of the AAAI conference on artificial intelligence}, vol.~35, no.~3, 2021, pp. 2458--2466.

\bibitem{zou2023object}
Z.~Zou, K.~Chen, Z.~Shi, Y.~Guo, and J.~Ye, ``Object detection in 20 years: A survey,'' \emph{Proceedings of the IEEE}, 2023.

\bibitem{he2017mask}
K.~He, G.~Gkioxari, P.~Doll{\'a}r, and R.~Girshick, ``Mask r-cnn,'' in \emph{Proceedings of the IEEE international conference on computer vision}, 2017, pp. 2961--2969.

\bibitem{sun2021sparsercnn}
P.~Sun, R.~Zhang, Y.~Jiang, T.~Kong, C.~Xu, W.~Zhan, M.~Tomizuka, L.~Li, Z.~Yuan, C.~Wang \emph{et~al.}, ``Sparse r-cnn: End-to-end object detection with learnable proposals,'' in \emph{Proceedings of the IEEE/CVF conference on computer vision and pattern recognition}, 2021, pp. 14\,454--14\,463.

\bibitem{Cheng2022DOD}
G.~Cheng, Y.~Yao, S.~Li, K.~Li, X.~Xie, J.~Wang, X.~Yao, and J.~Han, ``Dual-aligned oriented detector,'' \emph{IEEE Transactions on Geoscience and Remote Sensing}, vol.~60, pp. 1--11, 2022.

\bibitem{Yao2023OIBBR}
Y.~Yao, G.~Cheng, G.~Wang, S.~Li, P.~Zhou, X.~Xie, and J.~Han, ``On improving bounding box representations for oriented object detection,'' \emph{IEEE Transactions on Geoscience and Remote Sensing}, vol.~61, pp. 1--11, 2023.

\bibitem{zhou2019ExtremeNet}
X.~Zhou, J.~Zhuo, and P.~Krahenbuhl, ``Bottom-up object detection by grouping extreme and center points,'' in \emph{Proceedings of the IEEE/CVF conference on computer vision and pattern recognition}, 2019, pp. 850--859.

\bibitem{tian2019fcos}
Z.~Tian, C.~Shen, H.~Chen, and T.~He, ``Fcos: Fully convolutional one-stage object detection,'' in \emph{Proceedings of the IEEE/CVF international conference on computer vision}, 2019, pp. 9627--9636.

\bibitem{Cheng2023SFRNet}
G.~Cheng, Q.~Li, G.~Wang, X.~Xie, L.~Min, and J.~Han, ``Sfrnet: Fine-grained oriented object recognition via separate feature refinement,'' \emph{IEEE Transactions on Geoscience and Remote Sensing}, vol.~61, pp. 1--10, 2023.

\bibitem{Cheng2022AOPG}
G.~Cheng, J.~Wang, K.~Li, X.~Xie, C.~Lang, Y.~Yao, and J.~Han, ``Anchor-free oriented proposal generator for object detection,'' \emph{IEEE Transactions on Geoscience and Remote Sensing}, vol.~60, pp. 1--11, 2022.

\bibitem{Wang2022GFL}
J.~Wang, F.~Li, and H.~Bi, ``Gaussian focal loss: Learning distribution polarized angle prediction for rotated object detection in aerial images,'' \emph{IEEE Transactions on Geoscience and Remote Sensing}, vol.~60, pp. 1--13, 2022.

\bibitem{yang2021r3det}
X.~Yang, J.~Yan, Z.~Feng, and T.~He, ``R3det: Refined single-stage detector with feature refinement for rotating object,'' in \emph{Proceedings of the AAAI conference on artificial intelligence}, vol.~35, no.~4, 2021, pp. 3163--3171.

\bibitem{mardia2000directional}
K.~V. Mardia, P.~E. Jupp, and K.~Mardia, \emph{Directional statistics}.\hskip 1em plus 0.5em minus 0.4em\relax Wiley Online Library, 2000, vol.~2.

\bibitem{ding2021DOTA}
J.~Ding, N.~Xue, G.-S. Xia, X.~Bai, W.~Yang, M.~Y. Yang, S.~Belongie, J.~Luo, M.~Datcu, M.~Pelillo \emph{et~al.}, ``Object detection in aerial images: A large-scale benchmark and challenges,'' \emph{IEEE transactions on pattern analysis and machine intelligence}, vol.~44, no.~11, pp. 7778--7796, 2021.

\bibitem{Liu2017HRSC2016}
Z.~Liu, L.~Yuan, L.~Weng, and Y.~Yang, ``A high resolution optical satellite image dataset for ship recognition and some new baselines,'' in \emph{International Conference on Pattern Recognition Applications and Methods}, 2017.

\bibitem{lin2017focal}
T.-Y. Lin, P.~Goyal, R.~Girshick, K.~He, and P.~Doll{\'a}r, ``Focal loss for dense object detection,'' in \emph{Proceedings of the IEEE international conference on computer vision}, 2017, pp. 2980--2988.

\bibitem{wang2019maskobb}
J.~Wang, J.~Ding, H.~Guo, W.~Cheng, T.~Pan, and W.~Yang, ``Mask obb: A semantic attention-based mask oriented bounding box representation for multi-category object detection in aerial images,'' \emph{Remote Sensing}, vol.~11, no.~24, p. 2930, 2019.

\bibitem{wang2020CenterMap}
J.~Wang, W.~Yang, H.-C. Li, H.~Zhang, and G.-S. Xia, ``Learning center probability map for detecting objects in aerial images,'' \emph{IEEE Transactions on Geoscience and Remote Sensing}, vol.~59, no.~5, pp. 4307--4323, 2020.

\bibitem{Rao2023FCOSF}
C.~Rao, J.~Wang, G.~Cheng, X.~Xie, and J.~Han, ``Learning orientation-aware distances for oriented object detection,'' \emph{IEEE Transactions on Geoscience and Remote Sensing}, vol.~61, pp. 1--11, 2023.

\bibitem{yang2022scrdet++}
X.~Yang, J.~Yan, W.~Liao, X.~Yang, J.~Tang, and T.~He, ``Scrdet++: Detecting small, cluttered and rotated objects via instance-level feature denoising and rotation loss smoothing,'' \emph{IEEE Transactions on Pattern Analysis and Machine Intelligence}, vol.~45, no.~2, pp. 2384--2399, 2022.

\bibitem{he2016resnet}
K.~He, X.~Zhang, S.~Ren, and J.~Sun, ``Deep residual learning for image recognition,'' in \emph{Proceedings of the IEEE conference on computer vision and pattern recognition}, 2016, pp. 770--778.

\bibitem{lin2017fpn}
T.-Y. Lin, P.~Doll{\'a}r, R.~Girshick, K.~He, B.~Hariharan, and S.~Belongie, ``Feature pyramid networks for object detection,'' in \emph{Proceedings of the IEEE conference on computer vision and pattern recognition}, 2017, pp. 2117--2125.

\bibitem{lin2014coco}
T.-Y. Lin, M.~Maire, S.~Belongie, J.~Hays, P.~Perona, D.~Ramanan, P.~Doll{\'a}r, and C.~L. Zitnick, ``Microsoft coco: Common objects in context,'' in \emph{Computer Vision--ECCV 2014: 13th European Conference, Zurich, Switzerland, September 6-12, 2014, Proceedings, Part V 13}.\hskip 1em plus 0.5em minus 0.4em\relax Springer, 2014, pp. 740--755.

\bibitem{everingham2009pascal}
M.~Everingham, ``The pascal visual object classes challenge 2007,'' in \emph{http://www. pascal-network. org/challenges/VOC/voc2007/workshop/index. html}, 2009.

\bibitem{ming2021RIDet}
Q.~Ming, L.~Miao, Z.~Zhou, X.~Yang, and Y.~Dong, ``Optimization for arbitrary-oriented object detection via representation invariance loss,'' \emph{IEEE Geoscience and Remote Sensing Letters}, vol.~19, pp. 1--5, 2021.

\bibitem{liu2021OBD}
Y.~Liu, T.~He, H.~Chen, X.~Wang, C.~Luo, S.~Zhang, C.~Shen, and L.~Jin, ``Exploring the capacity of an orderless box discretization network for multi-orientation scene text detection,'' \emph{International Journal of Computer Vision}, vol. 129, pp. 1972--1992, 2021.

\end{thebibliography}
}

\begin{IEEEbiography}[{\includegraphics[width=1in,height=1.25in,clip,keepaspectratio]{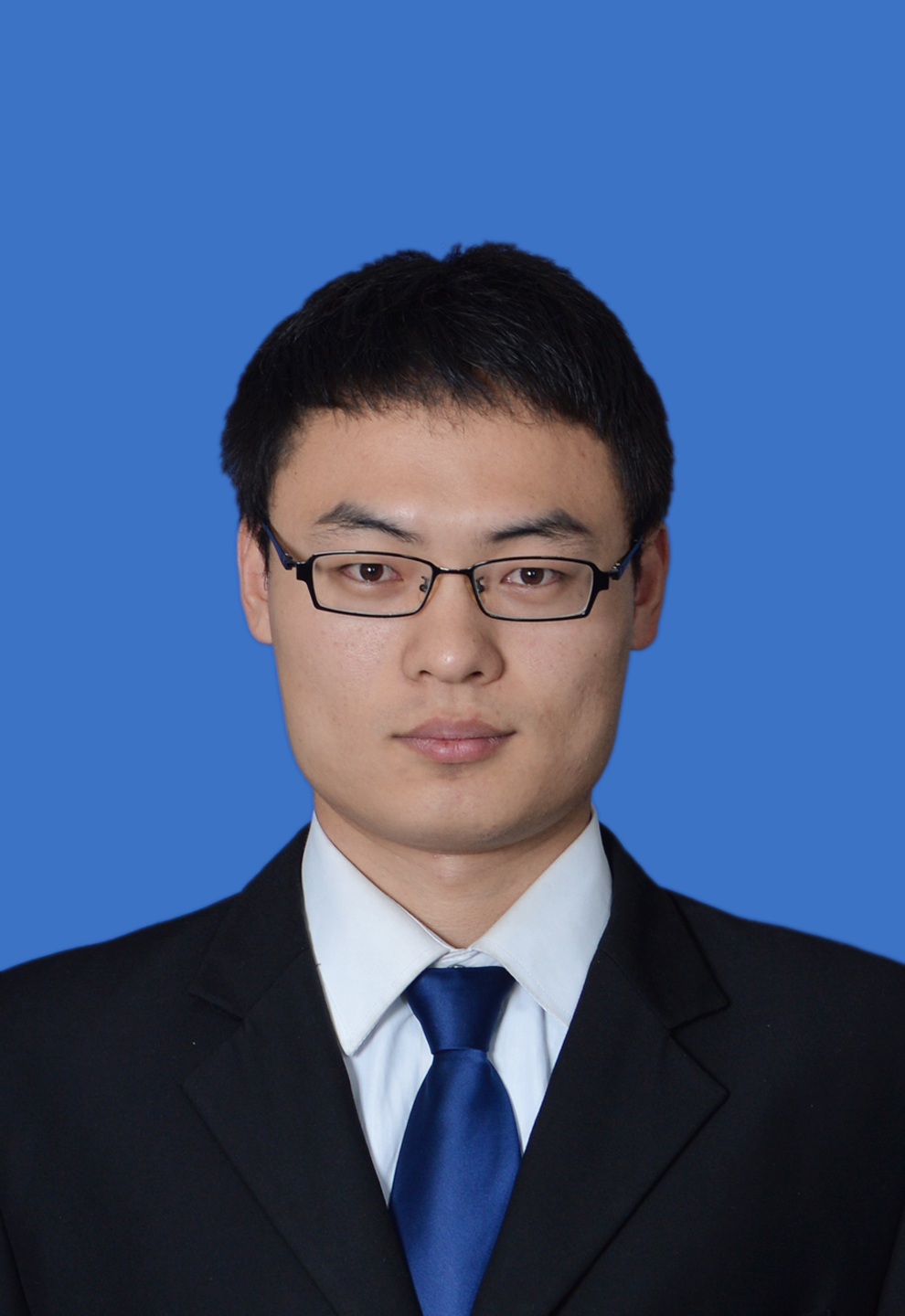}}]{Zifei Zhao}
received the M.S. and the B.E degree in photogrammetry and remote sensing from Shandong University of Science and Technology, Qingdao, China, 2015 and 2018. He is currently pursuing the Ph.D. degree in Technology and Engineering Center for Space Utilization, Chinese Academy of Sciences (CAS), Beijing, China.

His research interests include remote sensing image processing and satellite video analysis, such as object detection.
\end{IEEEbiography}

\begin{IEEEbiography}[{\includegraphics[width=1in,height=1.25in,clip,keepaspectratio]{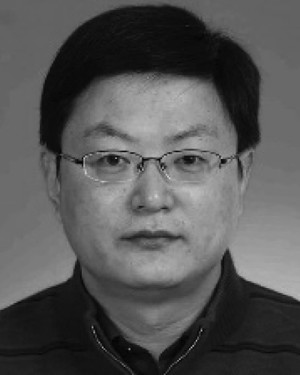}}]{Shengyang Li}
Technology and Engineering Center for Space Utilization, Chinese Academy of Sciences, Beijing, China
Key Laboratory of Space Utilization, Chinese Academy of Sciences, Beijing, China
University of Chinese Academy of Sciences, Beijing, China
Shengyang Li received the Ph.D. degree from the Institute of Remote Sensing Applications, Chinese Academy of Sciences, Beijing, China, in 2006.

He is currently a Professor with the Technology and Engineering Center for Space Utilization, Chinese Academy of Sciences. His research activities are machine learning in remote sensing image interpretation, deep learning in satellite video processing and analysis, intelligent image processing, analysis and understanding for space utilization, and space scientific big data modeling and analysis.
\end{IEEEbiography}

\vspace{11pt}

\vfill

\end{document}